\documentclass{article}

\usepackage[final, nonatbib]{tccml_neurips_2020}

\usepackage[utf8]{inputenc} 
\usepackage[T1]{fontenc}    
\usepackage{hyperref}       
\usepackage{url}            
\usepackage{booktabs}       
\usepackage{amsfonts}       
\usepackage{nicefrac}       
\usepackage{microtype}      
\usepackage{graphicx}

\title{Analyzing Sustainability Reports Using Natural Language Processing}

\author{%
Alexandra (Sasha) Luccioni\\
Université de Montréal + Mila \\
  \And
  Emily (Emi) Baylor\\
  McGill University\\
  \AND
 Nicolas Duchene \\
 Université de Montréal\\
}

\begin{document}

\maketitle

\begin{abstract}
Climate change is a far-reaching, global phenomenon that will impact many aspects of our society, including the global stock market~\cite{dietz2016climate}. In recent years, companies have increasingly been aiming to both mitigate their environmental impact and adapt to the changing climate context. This is reported via increasingly exhaustive reports, which cover many types of climate risks and exposures under the umbrella of Environmental, Social, and Governance (ESG). However, given this abundance of data, sustainability analysts are obliged to comb through hundreds of pages of reports in order to find relevant information. We leveraged recent progress in Natural Language Processing (NLP) to create a custom model, ClimateQA, which allows the analysis of financial reports in order to identify climate-relevant sections based on a question answering approach. We present this tool and the methodology that we used to develop it in the present article.
\end{abstract}

\section{Introduction}
In the coming years and decades, climate change will have a major impact on many global systems and structures, from agriculture to transportation and urban planning, affecting individual and collective behavior~\cite{rolnick2019tackling}. Its impact on the global stock market alone will be extensive, with damages estimated to be in the trillions of dollars~\cite{dietz2016climate}. However, it is difficult to predict exactly how and where climate change will impact financial assets, largely due to the lack of quantitative data on the subject. Nonetheless, gathering data regarding the risks and exposure that climate change poses to specific companies, as well as organizations' efforts to address and mitigate this risk, is a key part of this effort of predicting the extent of climate change impacts on the stock market. This data is often in textual format, buried in hundreds of pages of financial documents which must be manually analyzed, requiring significant time and effort.

Disclosing climate change risks and liabilities currently consists of a mix of mandatory and voluntary initiatives, resulting in disclosures that are often very heterogeneous and lack structure with regards to the subjects they cover, the metrics they use and the extent to which climate risk is quantified. 
To improve the state of voluntary climate disclosing and to encourage companies to increase their climate transparency, the Task Force on Climate-related Financial Disclosures (TCFD) was founded in 2015. In 2017, the TCFD released a set of recommendations to help structure and formalize companies' sustainability and climate risk reporting~\cite{tcfd2017final}. One of the key proposals that they made was a set of 14 questions to guide sustainability reporting, covering many different topics, from quantifying greenhouse gas emissions to ensuring the proper identification of climate-related risks and/or opportunities by the board of directors. In recent years, the TCFD questions are extensively used to guide the analysis of climate risk disclosures, with analysts using them to assess the extent and type of climate exposure of companies. 

The TCFD itself also uses the 14 questions to publish their yearly status reports on the progress of sustainability reporting. In 2019, they estimated that out of the 1000 companies whose reports they analyzed, only 29\% made relevant climate disclosures, stating that they were \emph{``concerned that not enough companies are disclosing decision-useful climate-related financial information"}~\cite{tcfd2019}. However, this figure is hard to calculate since it is based on a subset of the total number of companies and since the current approach used to analyse the reports is based on keyword search for analyst-defined terms. This suggests that much of the available data regarding financial climate disclosures is under-utilized or simply ignored.  In recent years, a few proposals have also been made regarding using NLP for analyzing sustainability reporting, signifying that this field is gathering momentum~\cite{luccioni2019using,becker2020}. The goal of our research project is to use Natural Language Processing (NLP) to create a tool allowing more efficient analysis of financial reports, reducing the time and effort required to identify climate-relevant disclosures.


\section{Model and Approach}

 Transformer models~\cite{vaswani2017attention} have had substantial success in many NLP tasks, from question answering~\cite{lan2019albert} to machine translation~\cite{ott2018scaling} and natural language inference (NLI)~\cite{conneau2017supervised}. In recent years, different variations of original architectures such as BERT~\cite{devlin2018bert} have also been used for domain-specific applications, from biomedical text mining~\cite{lee2020biobert} to sentiment analysis~\cite{mohammad2016sentiment}, and have already been trained on financial documents~\cite{araci2019finbert}. We adopted this Transformer-based approach to develop ClimateQA, our tool for extracting climate-relevant passages in financial documents; we will describe our approach in more detail below.

\subsection{Pretraining on Unlabelled Data}

Progress in NLP applications in finance has proven to be challenging notably because of the specialized language used: terms such as `bull' and `short' do not have the same meanings in finance as in general discourse~\footnote{A `bull' market is one that is on the rise; to `short' a stock means investing in such a way that the investor will profit if the value of the stock falls.} , whereas technical terms such as `liquidity' or `Keynesian' may not even be present in training corpora. In fact, research in financial NLP has found that using general-purpose NLP models trained on corpora such as Wikipedia and the Common Crawl fail to capture domain-specific terms and concepts which are critical for a coherent representation of the financial lexicon, and are therefore difficult to use out-of-the-box for financial tasks~\cite{araci2019finbert}.

To address this issue, we scraped 2,249 publicly available financial and sustainability reports from sources such as \underline{\href{https://www.sec.gov/edgar.shtml}{EDGAR}} and the \underline{\href{https://database.globalreporting.org/}{Global Reporting Initiative}} databases. These documents covered over 10 years of reports from a variety of publicly-traded companies, in sectors ranging from banking to agriculture. We extracted the raw text from the PDFs of the reports using the \underline{\href{https://github.com/chrismattmann/tika-python}{Tika package}} and used the raw text output to pre-train a word embedding model on the documents. Our hope in doing so was to reflect the context-specific nature of financial discourse, and to better represent the vocabulary used in sustainability reports.

\subsection{Fine-tuning on Labeled Data}

 Given the prevalence of the TCFD questions and the diversity of subjects that they cover, we used them to guide the analysis of sustainability reports. More specifically, we framed our task as one of question answering: given one of the 14 TCFD questions and a set of candidate sentences extracted from a financial report, we trained a model to determine whether or not a given sentence is an answer to one of the questions (see Table~\ref{qna} for examples of TCFD questions and answers). 
 
  \begin{table}[h!]
\caption{Examples of Question-Answer pairs from our corpus}
\label{qna}
  \centering
\begin{tabular}{|l|l|}
\hline
\textbf{TCFD Question} &
  \textbf{Answer Passage} \\ \hline
\begin{tabular}[c]{@{}l@{}}Does the organization describe the board's\\  (or board committee's) oversight of \\ climate-related risks and/or opportunities?\end{tabular} &
  \textit{\begin{tabular}[c]{@{}l@{}}The Company’s Audit Committee has the delegated \\ risk management oversight responsibility and \\ receives updates on the risk management processes \\ and key risk factors on a quarterly basis.\end{tabular}} \\ \hline
\begin{tabular}[c]{@{}l@{}}Does the organization describe the\\  climate-related risks or opportunities \\ the organization has identified?\end{tabular} &
  \textit{\begin{tabular}[c]{@{}l@{}} The availability and price of these commodities are \\subject to factors such as changes in weather \\ conditions, plantings, and government policies\end{tabular}} \\ \hline
\end{tabular}
\end{table}

In order to gather labeled data, we reached out to a team of sustainability analysts, who were able to provide us with a small set of financial reports from previous years, hand-labeled using the 14 TCFD questions. Based on these reports, we constructed our training set for the question answering task: positive examples consisted of pairs of questions and sentences which contained the answers to the questions, whereas negative examples were generated by pairing the remaining sentences with the questions that they did not answer. 
We split this randomly by company into three sets (training, development and test) and took stratified random samples of each set, separating the documents on a per-company basis. Our ClimateQA model was therefore trained on 15,000 negative examples and 1,500 positive examples, whereas the development set comprised of 7,500 negative examples and 750 positive examples, while the test set had 1,200 negative and 400 positive examples.

 
\section{Results}

\subsection{Comparing Large and Base Models}

To ensure the best performance of our model, we decided to use the RoBERTa (Robustly optimized BERT approach) architecture, whose performance was found to be either matching or exceeding that of the original BERT architecture on a variety of tasks~\cite{liu2019roberta}, trained on a masked language model task on data from Wikipedia and the Common Crawl. However, there are in fact two different versions of the RoBERTa architecture: RoBERTa-Large, which has 355M parameters, and RoBERTa-Base, which has only 125M. In traditional NLP applications, RoBERTa-Large is often the preferred version of BERT due to it being considered larger and more accurate~\cite{liu2019roberta}. However, more parameters comes with significantly higher memory requirements and a longer training time -- in our case, RoBERTa-Large took almost 12 hours to train on a 12 GB GPU, whereas RoBERTa-Base took less than 5 hours. We therefore compared the performance of both models based on their F1 score, given the large class imbalance of our data. A metric we were also particularly interested in was the difference between the validation and test scores, since this is a good indicator of our model's capacity to generalize. As can be seen in Table~\ref{compare} while RoBERTa-Large does slightly better than RoBERTa-Base across the board, the differences are minor, between 0.5 and 2.5\%. Given the major difference between the two models in terms of memory constraints and training time, and the importance of energy efficiency in choosing neural network architectures~\cite{lacoste2019quantifying}, we decided to use RoBERTa-Base for further hyper-parameter tuning and online deployment.

\begin{table}[h!]
\caption{Comparing RoBERTa-Large and RoBERTa-Base}
\label{compare}
\begin{tabular}{l|l|l|l|l}
           & Train F1 & Validation F1 & Test F1 & \begin{tabular}[c]{@{}l@{}}Val-Test \\ Difference\end{tabular} \\ \hline
RoBERTa-Large &   99.9\%   & 92.2\%        & 85.5\%  & -6.7\%                                                          \\ \hline
RoBERTa-Base  &   99.9\%   & 91.7\%        & 82\%    & -9.7\%                                                         
\end{tabular}
\end{table}

\subsection{Analyzing Results by Sector and by Question}

Both the labelled and unlabelled datasets that we trained ClimateQA on were from a variety of sectors, which made them differ slightly in the terminology used and the types of disclosures made. For instance, the climate change-related risks identified by insurance companies was mostly due to physical risk (e.g. due to coastal properties being damaged by repeated flooding), whereas those identified in the energy sector can be linked to market or legislative risk (e.g. for oil and gas companies impacted by higher taxation rates or customers switching to renewable energy). We therefore analyzed our results by sector to get a better idea of which sectors had the best performance. We found that the Energy sector had the best results, most likely due to the homogeneity of the companies in the labeled data we received -- most of it was from oil and gas companies who disclosed very similar risks and opportunities and often used extensive boilerplate language. Overall, the generalization capacity of ClimateQA in respect to the different sectors was good across the board, with an average 13.3\% difference between validation and test scores, which is to be expected since they consisted of different companies.

\begin{table}[h!]
  \centering
  \caption{ClimateQA results by sector}
\begin{tabular}{l|l|l|l}

 &
  \begin{tabular}[c]{@{}l@{}}Validation\\ F1 Score\end{tabular} &
  \begin{tabular}[c]{@{}l@{}}Test F1\\ Score\end{tabular} &
  \begin{tabular}[c]{@{}l@{}}Val - Test \\ Difference\end{tabular} \\ \hline
Agriculture, Food \& Forests & 89.4\% & 72.1\% & -17.2\% \\ \hline
Energy                       & 94.2\% & 89.8\% & -4.4\% \\ \hline
Banks                        & 91.9\% & 86.6\% & -5.3\% \\ \hline
Transportation               & 86.9\% & 72.5\% & -14.4\% \\ \hline
Insurance                    & 92.9\% & 78.7\% & -14.2\% \\ \hline
Materials \& Buildings       & 91.8\% & 67.6\% & -24.2\% \\ \hline \textbf{Average across sectors}            & \textbf{91.7\%} & \textbf{82.0\%} & \textbf{-9.7\%}
\end{tabular}
\end{table}

We also looked at ClimateQA's performance on each of the 14 TCFD questions (a detailed per-question analysis is presented in Appendix 1). This comparison is difficult to do systematically, because some questions are answered much more frequently than others. For instance, Question 1 (\emph{``Does the organization describe the board's  oversight of climate-related risks  and/or opportunities?"}) is answered by all companies in our dataset, whereas Question 12 (\emph{``Does the organization disclose Scope 1 and Scope 2, and, if appropriate Scope 3 GHG emissions?"}) is answered much less frequently, with only 8\% of companies providing answers. Nonetheless, there are some observations can be made, for instance the fact that Question 4 (\emph{``Does the organization describe time frames associated with its climate-related risks or opportunities?"}) shows the worst performance; we believe that this is due to its genericity, since time frames can be anything from numbers, e.g. `by 2025', to time horizons, e.g. `within the next five years. Furthermore, Question 10 (\emph{``Does the organization describe how processes for identifying, assessing, and managing climate-related risks are integrated into the organization's overall risk management?"}) has the most significant differences between validation and test data, up to 51\%. We believe that this is due to the specificity and diversity of answers, since these processes that may vary very much depending on the size and type of company that is involved. Nonetheless, it is interesting to observe which questions were easier to generalize for the ClimateQA model, and which needed more data to attain better performance. 

\section{The Final Tool and Next Steps}
While the NLP research involved in our project was an interesting and worthwhile endeavor in itself, our end goal was always to create a user-friendly sustainability report tool that could easily be used by analysts. This is why we have spent a significant amount of time and effort deploying our ClimateQA model. To this end, the model is hosted on the Microsoft Azure cloud solution (see Appendix 2 for high-level architecture), allowing users to interact with a web application without needing ML expertise. Via the website, a user is able to upload PDF files to be analyzed and receive a batch ID which they can subsequently use to check if they have been processed. In terms of the processing pipeline, once the PDF files are uploaded by the user, they are sent to an Azure blob storage; this triggers an Azure ML pipeline composed of three steps: (1) Extraction of the raw text from the PDF, (2) Parsing and splitting the text into a TSV file, and (3) Inference using the ClimateQA model, which identifies sections of the text that answer the question(s) submitted. The results are sent to the blob storage, and the user is able to download the results of the inference in TSV format and use them to support their analysis.

We are working on two future directions for our project: improving the model itself and improving the user experience of the website. On the one hand, error analysis of our results has brought to light that sentences that were part of a PDF table were often not identified by our approach. We are therefore currently working on improving our text extraction approach, notably by exploring commercial PDF extraction documents to identify rows and cells within tables in the reports themselves. We are also looking into ways of better utilizing the pre-trained word embedding models in our approach; while our current model does well now, we believe that this would help it work even better on the finance-specific vocabulary that is present in the reports. On the other hand, we are working on the website so that it can display results interactively, allowing users to visualize the passages identified by the model within the documents. This visualization tool will help democratize ClimateQA by bridging the gap between end-users and a specialized climate finance model, all the while keeping a simple user interface abstracting away all implementation details.


\newpage
\bibliographystyle{unsrt}
\bibliography{bibliography}
\newpage
\section*{Appendix 1} \label{appendix1}

\begin{table}[h!]

  \caption{ClimateQA results by question}
\begin{tabular}{l|l|l|l}
\begin{tabular}[c]{@{}l@{}}TCFD Question\end{tabular} &
  \begin{tabular}[c]{@{}l@{}}Validation\\  F1 Score\end{tabular} &
  \begin{tabular}[c]{@{}l@{}}Testing\\  F1 Score\end{tabular} &
  \begin{tabular}[c]{@{}l@{}}Val - Test \\ Difference\end{tabular} \\ \hline
\begin{tabular}[c]{@{}l@{}}1) Does the organization describe the board's  oversight of \\ climate-related risks  and / or opportunities?\end{tabular} &
  97.78\% &
  84.85\% &
  -12.93\% \\ \hline
\begin{tabular}[c]{@{}l@{}}2) Does the organization describe management's role in \\ assessing and managing climate-related risks and/or opportunities?\end{tabular} &
  96.60\% &
  84.75\% &
  -11.85\% \\ \hline
\begin{tabular}[c]{@{}l@{}}3) Does the organization describe the climate-related risks or\\ opportunities the organization has identified?\end{tabular} &
  90.61\% &
  89.50\% &
  -1.11\% \\ \hline
\begin{tabular}[c]{@{}l@{}}4) Does the organization describe time frames (short, medium, or \\ long term) associated with its climate-related risks or opportunities?\end{tabular} &
  75.00\% &
  N/A &
  N/A \\ \hline
\begin{tabular}[c]{@{}l@{}}5) Does the organization describe the impact of climate-related risks \\ and opportunities on the organization?\end{tabular} &
  90.91\% &
  86.59\% &
  -4.32\% \\ \hline
\begin{tabular}[c]{@{}l@{}}6) Does the organization describe the resilience of its strategy, taking\\ into consideration different climate-related scenarios, including a \\ potential future state aligned with the Paris Agreement?\end{tabular} &
  94.12\% &
  N/A &
  N/A \\ \hline
\begin{tabular}[c]{@{}l@{}}7) Does the organization disclose the use of a 2C scenario in evaluating \\ strategy or financial planning, or for other business purposes?\end{tabular} &
  100.00\% &
  100.00\% &
  0.00\% \\ \hline
\begin{tabular}[c]{@{}l@{}}8) Does the organization describe the organization's processes for \\ identifying and/or assessing climate-related risks?\end{tabular} &
  89.87\% &
  81.08\% &
  -8.79\% \\ \hline
\begin{tabular}[c]{@{}l@{}}9) Does the organization describe the organization's processes for \\ managing climate-related risks?\end{tabular} &
  92.54\% &
  60.00\% &
  -32.54\% \\ \cline{1-3}
\begin{tabular}[c]{@{}l@{}}10) Does the organization describe how processes for identifying, \\ assessing, and managing climate-related risks are integrated into \\ the organization's overall risk management?\end{tabular} &
  96.15\% &
  44.44\% &
  -51.71\% \\ \hline
\begin{tabular}[c]{@{}l@{}}11) Does the organization disclose the metrics it uses to assess climate-\\ related risks and/or opportunities?"\end{tabular} &
  88.48\% &
  90.67\% &
  2.18\% \\ \hline
\begin{tabular}[c]{@{}l@{}}12) Does the organization disclose Scope 1 and Scope 2, and, if \\ appropriate Scope 3 GHG emissions?\end{tabular} &
  97.50\% &
  94.74\% &
  -2.76\% \\ \hline
\begin{tabular}[c]{@{}l@{}}13) Does the organization describe the targets it uses to manage climate-\\ related risks and/or opportunities?\end{tabular} &
  90.20\% &
  98.31\% &
  8.11\% \\ \hline
\begin{tabular}[c]{@{}l@{}}14) Does the organization describe its performance related to those \\ targets (referenced in question 13)?\end{tabular} &
  93.33\% &
  73.08\% &
  -20.26\% \\ \hline
Average &
  92.35\% &
  82.24\% &
  -10.98\%
\end{tabular}
\end{table}

\newpage



\end{document}